\documentclass[lettersize,journal]{IEEEtran}
\usepackage{amsmath,amsfonts}
\usepackage{algorithmic}
\usepackage{array}
\usepackage{textcomp}
\usepackage{stfloats}
\usepackage{csquotes}
\usepackage{url}
\usepackage{verbatim}
\usepackage{booktabs}

\usepackage{graphicx}
\usepackage{xcolor}
\usepackage{multirow}
\usepackage{subcaption} 

\def\BibTeX{{\rm B\kern-.05em{\sc i\kern-.025em b}\kern-.08em
    T\kern-.1667em\lower.7ex\hbox{E}\kern-.125emX}}
\usepackage{balance}
\begin{document}
\title{Enhancing Small Object Encoding in Deep Neural Networks: Introducing Fast\&Focused-Net with Volume-wise Dot Product Layer}
\author{Tofik~Ali, Partha~Pratim~Roy
\thanks{T. Ali and P.P. Roy are with the Department
of Computer Science and Engineering, Indian Institute of Technology Roorkee, India,
e-mail: tali@cs.iitr.ac.in, e-mail: proy.fcs@iitr.ac.in}
}

\maketitle

\begin{abstract}
In this paper, we introduce Fast\&Focused-Net, a novel deep neural network architecture tailored for efficiently encoding small objects into fixed-length feature vectors. Contrary to conventional Convolutional Neural Networks (CNNs), Fast\&Focused-Net employs a series of our newly proposed layer, the Volume-wise Dot Product (VDP) layer, designed to address several inherent limitations of CNNs. Specifically, CNNs often exhibit a smaller effective receptive field than their theoretical counterparts, limiting their vision span. Additionally, the initial layers in CNNs produce low-dimensional feature vectors, presenting a bottleneck for subsequent learning. Lastly, the computational overhead of CNNs, particularly in capturing diverse image regions by parameter sharing, is significantly high.
\par
The VDP layer, at the heart of Fast\&Focused-Net, aims to remedy these issues by efficiently covering the entire image patch information with reduced computational demand. Experimental results demonstrate the prowess of Fast\&Focused-Net in a variety of applications. For small object classification tasks, our network outperformed state-of-the-art methods on datasets such as CIFAR-10, CIFAR-100, STL-10, SVHN-Cropped, and Fashion-MNIST. In the context of larger image classification, when combined with a transformer encoder (ViT), Fast\&Focused-Net produced competitive results for OpenImages V6, ImageNet-1K, and Places365 datasets. Moreover, the same combination showcased unparalleled performance in text recognition tasks across SVT, IC15, SVTP, and HOST datasets. This paper presents the architecture, the underlying motivation, and extensive empirical evidence suggesting that Fast\&Focused-Net is a promising direction for efficient and focused deep learning.
\end{abstract}


\section{Introduction}
\label{Introduction}

Deep Neural Networks (DNNs) are a class of machine learning models that are designed to mimic the way the human brain works, with the aim of learning from large amounts of data. While a neural network with a single layer can still make approximate predictions, additional hidden layers can help optimize the accuracy. DNNs use this concept of 'depth', which refers to the number of hidden layers for which the learning process is repeated. This depth allows neural networks to learn through a hierarchical process of transforming the input data, layer by layer, to make more abstract and composite representations, thereby enabling the model to recognize complex patterns.

The focus of this paper is on the encoding of small objects into fixed length feature vectors by a deep neural network. This is a critical aspect of machine learning and computer vision, as it allows for the efficient and comprehensive representation of data, which in turn enables more accurate predictions and analyses. The encoding process involves the transformation of raw data into a format that can be easily processed by the neural network, and the quality of this encoding can significantly impact the performance of the model.

Despite the advancements in deep learning, there are still several challenges associated with the encoding of small objects in deep neural networks. One of the main issues is the effective receptive field of Convolutional Neural Networks (CNNs), which is often smaller than its theoretical receptive field. This means that the network is not able to fully capture the information in the input data~\cite{luo2016understanding},~\cite{liu2018understanding},~\cite{koutini2019receptive},~\cite{bruton2022translated}. Additionally, the initial layers of CNNs often generate a low-dimensional feature vector, which can limit the learning capacity of the subsequent layers. Furthermore, the computation cost of CNNs is high due to the need to generate feature vectors for different regions of the image through a sliding window and parameter-sharing approach.

To address these issues, we propose a new deep neural network, which we call Fast\&Focused-Net (FFN). Unlike traditional CNNs, FFN uses a new type of layer, called the Volume-wise Dot Product (VDP) layer, which requires less computation and covers the full image patch information. The VDP layer is designed to overcome the limitations of the convolution layer, by ensuring that the effective receptive field is equal to the theoretical receptive field, and by generating distinct feature vectors for all encountered image regions. This results in a more efficient and comprehensive encoding of small objects, thereby improving the performance of the deep neural network.

\section{Limitations of CNN}
\label{LimitationsofCNN}
Convolutional Neural Networks (CNNs) have been a cornerstone in the field of deep learning, particularly in image processing tasks. However, they are not without their limitations. This section will delve into the three main limitations of CNN-based networks as identified in the overview: the effective receptive field, the initial layer bottleneck, and the computation cost.

\subsection{Effective Receptive Field}

The receptive field of a CNN is the region in the input space that a particular CNN’s feature is looking at. The theoretical receptive field is computed based on the network's architecture and can be easily determined by calculating how much of the input the filters at each layer can see. However, the effective receptive field of a CNN-based network (stride is less than the kernel size) tends to be smaller than the theoretical receptive field~\cite{luo2016understanding}.

\begin{figure*}[ht]
    \centering
    \includegraphics[width=0.9\textwidth]{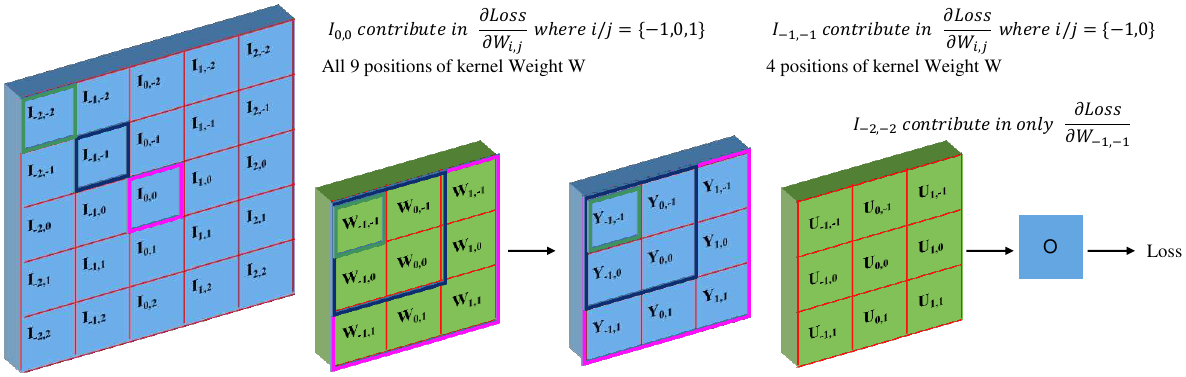}
    \caption{Illustration highlighting the influence of various regions within the input feature map on the gradient determination for the weights of a convolutional layer's kernel.}
    \label{fig:GradCalc}
\end{figure*}

The main reason for this discrepancy lies in the gradient calculation of the CNN kernel's weights in the training phase. The central area of the receptive field contributes more to the gradient calculation of CNN weights and updates the weights accordingly in the training phase (refer to Figure~\ref{fig:GradCalc}). As the training process progresses, the weights in the CNN layer increasingly favor the central region.

This phenomenon results in a CNN having a sharply defined vision at the center, while maintaining a less precise vision at the outer region. This allows it to gather the necessary context information, enabling superior decision-making within the focus region. However, this also means that a substantial amount of resources, both parameters and computational efforts, are invested in creating this less clear vision of the outer region, which could potentially be better optimized.

\subsection{Initial Layer Bottleneck}

The initial layers of a CNN are responsible for generating distinct vectors for all encountered image regions inside their receptive field. However, different image regions share the same parameter (same convolution layer) for the corresponding vector. This increases the total number of distinct image regions encountered by a convolution layer manyfold. 

The corresponding vector dimension is small (the first layer has generally a 64-dimensional output); therefore, learning a unique or distinguishable vector for these distinct image regions is hard to achieve. This limitation can potentially hinder the learning of the subsequent layers, creating a bottleneck in the network.

\subsection{Computation Cost}

The computation cost of a CNN network is very high, as it needs to generate feature vectors for different regions of the image through a sliding window and parameter-sharing approach. Although the initial layers of a CNN have a small number of kernels, they have to be applied across all regions of the input image, consuming a significant amount of overall computation.

Moreover, most of the receptive field of a CNN (near the final layer) is outside of the region of interest (ROI), which affects two very important aspects. First, the ROI is not paid full attention to by the layer, and second, the boundary regions are not covered properly (some padding element will be added instead of their actual neighbour in the large image that is cropped now). Collectively, CNN is doing a lot of calculations that are not needed to encode the information about ROI, leading to inefficient use of computational resources.

In summary, while CNNs have been instrumental in advancing the field of deep learning, these limitations highlight the need for more efficient and comprehensive encoding methods, such as the proposed Fast\&Focused-Net and VDP layer.

\section{Methodology}
\label{Methodology}
\subsection{The Volume-wise Dot Product (VDP) Layer}

The computation of the VDP layer is a straightforward process. The input feature map, which is the output from the previous layer, is initially divided into different hyper-volumes of the same dimensions. If we consider the input feature map to have the shape HxWxC, it can be divided into $(N_h, N_w, N_c)$ hyper-volumes of shape $(H/N_h, W/N_w, C/N_c)$. 

\begin{figure}[h!]
    \centering
    \begin{subfigure}[b]{0.45\textwidth} 
        \includegraphics[width=\textwidth]{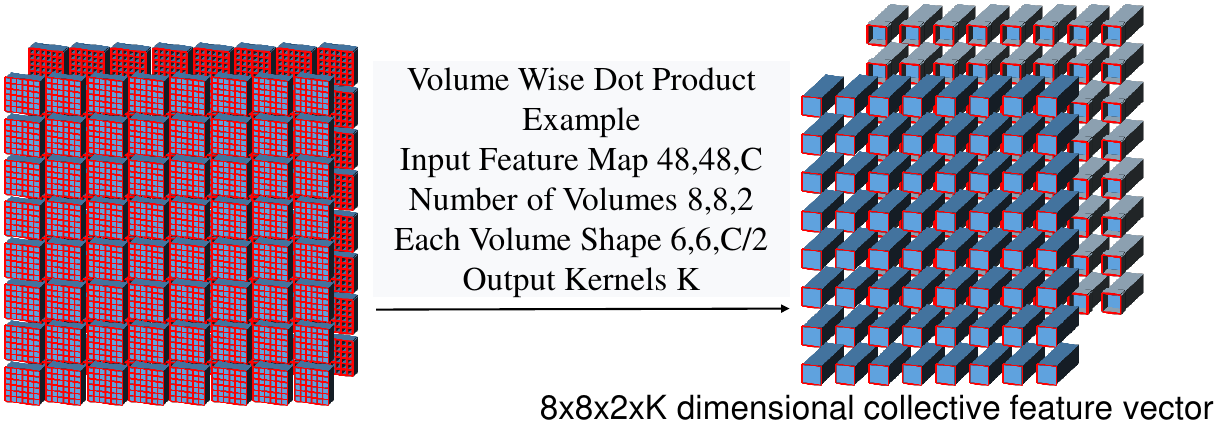}
        \caption{Visual illustration of the VDP layer computation.}
        \label{fig:first_image}
    \end{subfigure}
    \hfill 
    \begin{subfigure}[b]{0.45\textwidth}
        \includegraphics[width=\textwidth]{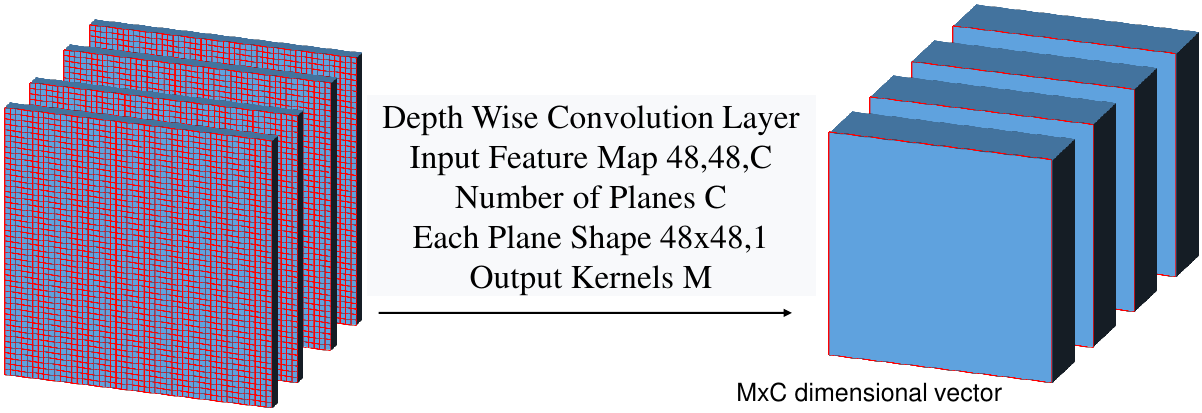}
        \caption{Schematic representation of the Depthwise Convolution layer.}
        \label{fig:DepthCNN}
    \end{subfigure}
    \caption{Comparative visualization of the Depthwise convolution layer and the VDP layer.}
    \label{fig:VDP_DepthCNN}
\end{figure}

Each of these volumes is then subjected to a dot product operation with its corresponding weight volume. Following this, a bias term is added, and an activation function is applied. This process is visually illustrated in Figure \ref{fig:VDP_DepthCNN} (a), where the VDP layer computation is compared with the depthwise convolution layer, as shown in Figure \ref{fig:VDP_DepthCNN} (b).

\begin{figure*}[h!]
    \centering
    \includegraphics[width=0.9\textwidth]{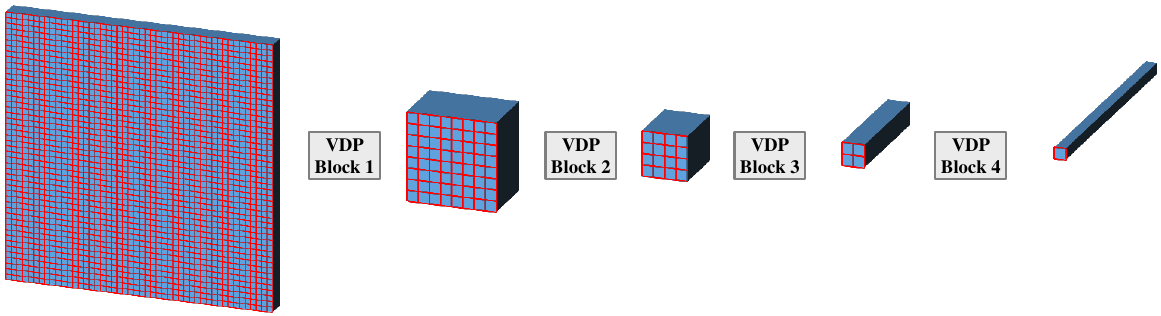} 
    \caption{Depiction of the Fast\&Focused-Net backbone encoding network, assembled by stacking VDP Blocks. Each block consists of several VDP layers. For a detailed breakdown of each VDP layer, refer to Table~\ref{tab:volume_shapes}.}
    \label{fig:FFN}
\end{figure*}

The VDP layer computation is designed to be less computationally intensive while covering the full image patch information. This is achieved by dividing the input image into non-overlapping smaller regions and processing them separately by different sets of parameters. As there is no parameter sharing the total number of parameter of VDP layer is high and each region in the input contribute to only its corresponding volume which creates a uniform gradient calculation. Thus each VDP layer covers the full image while maintaining the computation low.

\subsection{Structure of the Fast\&Focused-Net}

The Fast\&Focused-Net (FFN) is a deep neural network that is assembled by stacking VDP blocks. Each block consists of several VDP layers. The structure of the FFN is depicted in Figure \ref{fig:FFN}, where the backbone encoding network is shown. 

The FFN is designed to encode the underlying information and generate a final feature vector through a calculated stacking of the VDP layer with different $(N_h, N_w, N_c)$. This design allows the FFN to efficiently and comprehensively encode small objects into fixed-length feature vectors. 

The structure of the FFN is not based on CNN, which allows it to overcome some of the limitations associated with CNN-based networks. These include the smaller effective receptive field, the low-dimensional feature vector bottleneck in the initial layers, and the high computation cost.

\subsection{Handling of the Issues of CNN}

The design of the FFN and the VDP layer addresses the three main issues associated with CNN-based networks.

\textbf{Issue 1: Receptive Field} - The receptive field of the first layer of the FFN is the complete input patch, which means it does not include any regions outside of the region of interest (ROI). This ensures that every piece of data inside the ROI contributes equally to the final feature vector, leading to a full effective receptive field. This is a significant improvement over the CNN-based (stride is less than kernel size) networks, where the effective receptive field is always smaller than the theoretical receptive field.

\textbf{Issue 2: Initial Layer Bottleneck} - The initial layers of the FFN have a large number of parameters, which allows them to generate distinct feature vectors for all encountered image regions inside their receptive field. 
If we consider a single volume, it has a similar number of parameters as the CNN layer, but the CNN layer produces exactly the same response for the same region appearing at different locations in the patch, whereas it is not a constraint for the VDP layer as different regions (volumes) have different parameters. Therefore, similar regions in a patch can get different responses if they are helpful for the training target. Due to all these properties of VDP, it is less likely (possibilities are always there) that the initial layers are becoming bottlenecks for generating distinct feature vectors. This is unlike the initial layers of a CNN, which generate a low-dimensional feature vector that becomes a bottleneck for the learning of the subsequent layers.

\textbf{Issue 3: Computation Cost} - The computation cost of the FFN is significantly lower than that of a CNN network. This is because the equivalent stride of the VDP layer is the same as the volume's shape, and the volumes of a VDP layer are always significantly smaller than the input feature map. This reduces the number of computations required to encode the information about the ROI, making the FFN more efficient. This aspect of the VDP is influenced by the depthwise convolution layer~\cite{howard2017mobilenets} (refer Figure~\ref{fig:VDP_DepthCNN}). Please see Table~\ref{tab:MAC_compare} for a comparison of Fast\&Focused-Net output vector, number of parameters, and computation cost with other network architectures.

In summary, the FFN and the VDP layer provide an efficient and comprehensive solution for encoding small objects into fixed-length feature vectors, overcoming the limitations associated with CNN-based networks.

\setlength{\tabcolsep}{2pt}
\begin{table*}
\centering
\caption{Comparison of Different Network Designs for CIFAR-10 Image Classification Task. The ViT-L/16 is a 24-layered large model which uses patch size 16x16 as input, and ViT-B is a 12-layered base model. The proposed Fast\&Focused network does not share any parameter between different volumes; therefore, its number of parameters and MAC operations are the same.}
\label{tab:MAC_compare}
\begin{tabular}{|c|c|c|c|c|}
\hline
\multicolumn{5}{|c|}{Task: CIFAR-10 and Input Image Size: $32\times32\times3$ }\\
\multirow{2}{*}{Model} & \multirow{2}{*}{Parameters} & Activation & Output Vector & MAC \\
 &  & Layers & Length & Number \\
\hline
ViT-B/16~\cite{dosovitskiy2020image} & 86M & 12+1 & 768 & 324M \\
ViT-L/16~\cite{dosovitskiy2020image} & 307M & 24+1 & 1024 & 1153M \\
VGG8~\cite{nokland2019training} & 7.3M & 8 & 1024 & 905M \\
VGG11~\cite{nokland2019training} & 12M & 11 & 1024 & 1338M \\
Fast\&Focused & 12M & 8+1 & 2048 & 12M \\
\hline
\end{tabular}
\end{table*}

\section{Experimental Setup}
\label{ExperimentalSetup}
\begin{figure}[h!]
    \centering
    \begin{subfigure}[b]{0.45\textwidth} 
        \includegraphics[width=\textwidth]{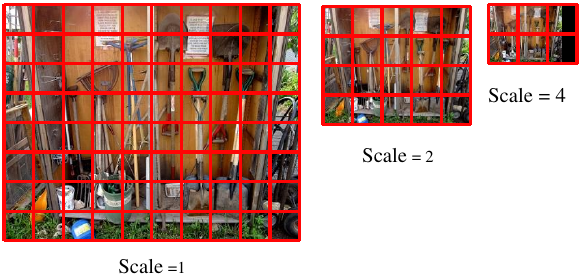}
        \caption{Multi-Scale Patching in Large Image Classification Task.}
        \label{fig: Multi-Scale in Large Image Classification}
    \end{subfigure}
    \hfill 
    \begin{subfigure}[b]{0.45\textwidth}
        \includegraphics[width=\textwidth]{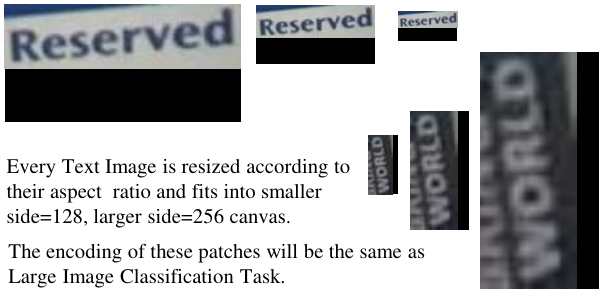}
        \caption{Multi-Scale Patching in Text Recognition Task.}
        \label{fig: Multi-Scale in Text Recognition}
    \end{subfigure}
    \caption{Visual illustration of Multi-Scale Patching.}
    \label{fig:multiscale}
\end{figure}

In all our experiments we are using only three different patch sizes 16x16, 32x32, and 96x96.
These patches are directly used for the small object classification task according to the dataset image sizes. If the size is not match then the images are resized to the nearest patch size. The large image classification and text recognition task require the division of the image into smaller patches, therefore these patches will have two coordinate positions so a 2D positional encoder will be used when processed with ViT. Besides this, Multi-Scale means that the patch size is 32x32, but the image is downscaled (2 and 4 times) to produce different patches. Then all these patches are passed to ViT-B for further processing with 3D positional encoding (Y, X, Scale) Scale will be 1,2,4 only~(refer Figure~\ref{fig: Multi-Scale in Large Image Classification}). For the text recognition task, every text Image is resized according to their aspect ratio and fits into smaller side=128, larger side=256 canvas. The encoding of these patches will be the same as Large Image Classification Task~(refer Figure~\ref{fig: Multi-Scale in Text Recognition}).

\setlength{\tabcolsep}{2pt}
\begin{table}[htbp]
\centering
\caption{The table presents a comprehensive comparison of several commonly used image classification datasets. For each dataset, the number of distinct classes and images allocated for training, validation, and testing purposes are provided. The datasets are categorized into 'Small Image Object Classification' and 'Large Image/Object Classification' as the evaluation tasks selected.}
\label{tab:dataset-details}
\begin{tabular}{|c|c|c|c|c|}
\hline
\multicolumn{5}{|c|}{Small Image Object Classification Dataset}\\
Dataset & Image & class & train & test\\
CIFAR-10 & 32x32x3 & 10 & 50000 & 10000\\
CIFAR-100 & 32x32x3 & 100 & 50000 & 10000\\
Fashion MNIST & 28x28x1 &	10 & 60000 & 10000\\
STL-10 & 96x96x3 & 10 & 5000 & 8000\\
SVHN - Cropped & 32x32x3 & 10 & 73257 & 26032\\
\hline
\hline
\multicolumn{5}{|c|}{Text Recognition Dataset}\\
Dataset & class & train & validation & test\\
SVT & & 257 & & 647\\
IC15 & & 4468 & & 2077\\
SVTP & & & & 645\\
HOST &  & 4832 & & 2416\\
\hline
\hline
\multicolumn{5}{|c|}{Large Image/Object Classification Dataset}\\
Dataset & class & train & validation & test\\
OpenImages V6 & 9600 & 9 million & 41620 & 125456\\
ImageNet & 1000 & 1.3 million & 50000 & 100000\\
Places365 & 365 & 1.8 million & 36000 & - \\
\hline
\end{tabular}
\end{table}

\setlength{\tabcolsep}{2pt}
\begin{table*}[h]
\centering
\caption{Details of the Fast\&Focused-Net architecture showing volume shapes, number of volumes, volume outputs, and output shapes for various colour image patch sizes: 16x16x3, 32x32x3, and 96x96x3.}
\label{tab:volume_shapes}
\begin{tabular}{|c|cccc|cccc|cccc|}
\hline
\multirow{3}{1cm}{VDP Layer} & \multicolumn{4}{|c|}{FFN for 16x16x3 image} & \multicolumn{4}{|c|}{FFN for 32x32x3 image} & \multicolumn{4}{|c|}{FFN for 96x96x3 image}\\
 & Volume & No. of & Volume & Output & Volume & No. of & Volume & Output & Volume & No. of & Volume & Output\\
 & Shape & Volumes & output & Shape & Shape & Volumes & output & Shape & Shape & Volumes & output & Shape\\
\hline
\hline
1 & 4x4x3 & 4x4x1 & 64 & 4x4x64 & 4x4x3 & 8x8x1 & 64 & 8x8x64 & 6x6x3 & 16x16x1 & 64 & 16x16x64 \\
2 & 1x1x64 & 4x4x1 & 64 & 4x4x64 & 1x1x64 & 8x8x1 & 64 & 8x8x64 & 1x1x64 & 16x16x1 & 64 & 16x16x64 \\
\hline
3 & 2x2x64 & 2x2x1 & 256 & 2x2x256 & 2x2x64 & 4x4x1 & 256 & 4x4x256 & 2x2x64 & 8x8x1 & 256 & 8x8x256 \\
4 & 1x1x256 & 2x2x1 & 256 & 2x2x256 & 1x1x256 & 4x4x1 & 256 & 4x4x256 & 1x1x256 & 8x8x1 & 256 & 8x8x256 \\
\hline
5 & 2x2x64 & 1x1x4 & 256 & 1x1x1024 & 2x2x64 & 2x2x4 & 256 & 2x2x1024 & 2x2x64 & 4x4x4 & 256 & 4x4x1024 \\
6 & 1x1x1024 & 1x1x1 & 1024 & 1x1x1024 & 1x1x1024 & 2x2x1 & 1024 & 2x2x1024 & 1x1x1024 & 4x4x1 & 1024 & 4x4x1024 \\
\hline
\hline
7 & - & - & - & - & 2x2x64 & 1x1x16 & 128 & 1x1x2048 & 2x2x64 & 2x2x16 & 128 & 2x2x2048 \\
8 & - & - & - & - & 1x1x2048 & 1x1x1 & 2048 & 1x1x2048 & 1x1x2048 & 2x2x1 & 2048 & 2x2x2048 \\
\hline
\hline
9 & - & - & - & - & - & - & - & - & 2x2x64 & 1x1x32 & 128 & 1x1x4096 \\
10 & - & - & - & - & - & - & - & - & 1x1x4096 & 1x1x1 & 4096 & 1x1x4096 \\
\hline
\end{tabular}
\end{table*}

\subsection{Tasks Description}

The performance of the proposed Fast\&Focused-Net (FFN) encoder network was evaluated using three distinct tasks: small image/object classification, large image classification, and text recognition.

\subsubsection{Small Image/Object Classification}

In this task, the FFN was used to directly classify small images or objects. The network encodes the information of the image patch into a final feature vector, which is then classified into respective classes through a linear transformation. This task is particularly useful in scenarios where the objects of interest are small and can be directly classified without the need for segmentation or localization.

\subsubsection{Large Image Classification}
In our approach for classifying large images, we employed the Fast\&Focused-Net (FFN) to initially encode smaller patches of these images. This was necessary because using FFN to encode a larger image directly would significantly increase the number of parameters, thereby elevating the risk of overfitting. To address this, we divided the large image into smaller patches and used FFN to encode each of these patches separately. After encoding, these vectors were processed by a 12-layer transformer attention-based model known as ViT-B, which is designed to handle 786-dimensional feature vectors. However, if the output feature vector from the FFN does not match this dimension, a linear transformation is required to align it with the ViT-B model's requirements. This transformation can potentially reduce the effectiveness of the FFN's feature vector, especially when there is a substantial difference in vector dimensions.

To overcome this limitation, we developed a new variant of the ViT model, which we named ViT-C4. This modified version of the ViT model consists of just 4 encoding layers, each equipped with a D-dimensional feature vector, a D-dimensional intermediate layer, and 16 multi-heads, where D represents the length of the output feature vector from the FFN. By aligning the dimensions of the ViT model with those of the FFN's output, ViT-C4 optimizes the utilization of the feature vectors, ensuring more effective and efficient image categorization.

\subsubsection{Text Recognition}

The text recognition task was similar to the large image classification task in that the large image was divided into smaller patches and encoded using the FFN. The encoded vectors were then processed by the ViT-B model to recognize the text. An ablation study was provided to show the impact of the patch size selection.

\subsection{Dataset Description}

The datasets used for the three tasks are described in Table \ref{tab:dataset-details}. For the small image/object classification task, the synthetic data used in the "Text Recognition" task was cropped at the character level and used for pretraining with the training set of all available real data. 
For the large image classification task, the imageNet-21K dataset with all other dataset's training sets is used in the pretraining of the underline model.

The pretraining for the text recognition task uses both synthetic and real datasets. Synthetic dataset is a combination of MJ~\cite{jaderberg2014synthetic} and ST~\cite{gupta2016synthetic}; and the Real dataset encompassing COCO, RCTW17, Uber, ArT, LSVT, MLT19, ReCTS, TextOCR, and OpenVINO datasets. Table~\ref{tab: results of text recognition task} show the different settings of the pretraining for a fair comparison. The abbreviations denote: S for Synthetic datasets; B represents Benchmark datasets SVT and IC15; and R indicates Real datasets.

\subsection{Handling Overfitting}

The proposed FFN has a large number of parameters for encoding a small region of interest, which puts the model at a high risk of overfitting. To handle this, we first pretrained the model for a large dataset of classes with large variations in the visual features of different classes. Then, the pretrained model was further fine-tuned with the respective dataset and task. The synthetic data and the training set of all available real data were used for the pretraining of the complete model (FFN + ViT-B). 

In addition to pretraining, we also utilized local learning~\cite{nokland2019training} during the pretraining of the model. Furthermore, we used dropout with a rate of 0.25 to regularize the model and prevent overfitting. Dropout is a technique where randomly selected neurons are ignored during training, which helps in preventing overfitting by reducing the co-adaptation of neurons.

\section{Results and Analysis}
\label{ResultsAnalysis}
The Fast\&Focused-Net (FFN) excels in small object classification tasks, primarily due to its capability to create extensive feature vectors and effectively utilize a large number of parameters. This design ensures that every pixel within its receptive field contributes equally, enhancing the model's performance in these specific tasks. However, when dealing with larger images, the model encounters challenges due to the increased total number of parameters, which often leads to overfitting. To address this, integration with other architectures becomes necessary. For instance, combining FFN with the Vision Transformer-C4 (ViT-C4), which matches FFN in feature dimensions, has shown superior performance. This hybrid approach manages to achieve this enhanced performance without significantly increasing computational and parametric demands, effectively leveraging the strengths of both architectures.

\setlength{\tabcolsep}{2pt}
\begin{table}[hbt]
\centering
\caption{Performance comparison of various methods on small object classification task across seven benchmark datasets. The top-performing method is highlighted in blue. Here FMN stands for Fashion MNIST and SVC stands for SVHN Cropped.}
\label{tab: results of small object classification task}
\begin{tabular}{|c|c|c|c|c|c|}
\hline
\multirow{2}{*}{Method} & \multicolumn{2}{|c|}{CIFAR} & FMN & STL & SVC\\
 & 10 & 100 &  & 10 & \\
\hline
ViT-L/16 &  99.42 & 93.90 & - & - & -\\
\cite{dosovitskiy2020image}& & & & & \\
\hline
Wide-ResNet101& 97.61 & 88.00 & - & 98.23 & 97.00\\
+ SpinalNet & & & & & \\
\cite{kabir2022spinalnet} & & & & & \\
\hline
VGG-5 + SpinalNet& 90.48 & 61.69 & 94.19 & 99.05 & 97.00\\
\cite{kabir2022spinalnet} & & & & & \\
\hline
WaveMix & 97.29 & 85.09 & - & - & 98.73\\
\cite{jeevan2022wavemix} & & & & & \\
\hline
ViT-L/16 & 99.05 & 93.3 & - & 99.71 & -\\
+ Spinal-FC & & & & & \\
+ Background & & & & & \\
\cite{kabir2023reduction} & & & & & \\
\hline
Fast\&Focused & \color{blue}99.48 & \color{blue}98.62 & \color{blue}98.19 & \color{blue}99.76 & \color{blue}98.87\\
\hline
\end{tabular}
\end{table}

\setlength{\tabcolsep}{2pt}
\begin{table}[hbt]
\centering
\caption{Performance comparison of various methods on text recognition task with a focus on word recognition accuracy across four benchmark datasets. The top-performing method is highlighted in blue. }
\begin{tabular}{|c|c|c|c|c|c|}
\hline
Method & Data & SVT & IC15 & SVTP & HOST \\
\hline
ASTER~ & S & 89.5 & 76.1 & 78.5 & – \\
\cite{shi2018aster}& & & & & \\
TextScanner & S* & 92.7 & - & 84.8 & – \\
\cite{wan2020textscanner}& & & & & \\
VisionLAN & S & 91.7 & 83.7 & 86.0 & 50.3 \\
\cite{wang2021two}& & & & & \\
ABINet & S & 93.5 & 86.0 & 89.3 & – \\
\cite{fang2021read}& & & & & \\
ViTSTR-B & S & 87.7 & 78.5 & 81.8 & – \\
\cite{atienza2021vision}& & & & & \\
PARSeq & S & 93.6 & 86.5 & 88.9 & – \\
\cite{bautista2022scene}& & & & & \\
CLIP4STR & S & 94.6 & 87.6 & 91.2 & 79.5 \\
\cite{zhao2023clip4str}& & & & & \\
TrOCR$_{Large}$ & S+B & 96.1 & 88.1 & 93.0 & – \\
\cite{li2023trocr}& & & & & \\
Fast\&Focused & S & 97.4 & 90.5 & 96.9 & 81.3 \\
Multi-Scale + ViT-B & & & & & \\
\hline
\hline
ViTSTR-S & R & 96.0 & 89.0 & 91.5 & 64.5 \\
\cite{atienza2021vision}& & & & & \\
ABINet & R & 98.2 & 90.5 & 94.1 & 72.2 \\
\cite{fang2021read}& & & & & \\
PARSeq & R & 97.9 & 90.7 & 95.7 & 74.4 \\
\cite{bautista2022scene}& & & & & \\
CLIP4STR & R & 98.3 & 91.4 & 97.2 & 77.5 \\
\cite{zhao2023clip4str}& & & & & \\
\hline
\hline
Fast\&Focused & R & \color{blue}98.6 & \color{blue}91.7 & \color{blue}97.5 & \color{blue}82.7 \\
Multi-Scale + ViT-B& & & & & \\
\hline
\end{tabular}
\label{tab: results of text recognition task}
\end{table}

\setlength{\tabcolsep}{2pt}
\begin{table}[htb]
\centering
\caption{Performance comparison of various methods on large image classification task across three benchmark datasets. The top-performing method is highlighted in blue. Here IN-1K is imageNet-1K, P-365 is Place365, OI-V6 is OpenImage V6, and NP is the number of parameters.}
\label{tab: results of large image classification task}
\begin{tabular}{|c|c|c|c|c|}
\hline
 & \multicolumn{3}{|c|}{Datasets} & NP \\
Method & IN-1K & P-365 & OI-V6 &  \\
& \multicolumn{2}{|c|}{Top 1 Accuracy} & mAP & \\
\hline
WaveMix-192/16 & 74.93 & 56.45 & - & 28.9 M \\
\cite{jeevan2022wavemix} & & & & \\
TResNet-M & - & - & 86.8 & - \\
\cite{ridnik2023ml} & & & & \\
P-ASL, Selective & - & - & 86.72 & - \\
\cite{ben2022multi} & & & & \\
TResNet-L & - & - & 87.34 & - \\
\cite{ben2022multi} & & & & \\
InternImage-T & 83.5 & - & - & 30 M \\
\cite{wang2023internimage} & & & & \\
InternImage-S & 84.2 & - & - & 50 M \\
\cite{wang2023internimage} & & & & \\
InternImage-B & 84.9 & - & - & 97 M \\
\cite{wang2023internimage} & & & & \\
ViT-L/16 + Lion& 85.59 & - & - & 305 M \\
\cite{singh2022revisiting} & & & & \\
ViT H/14& 88.6 & \color{blue}60.7 & - & 633.5 M \\
\cite{singh2022revisiting} & & & & \\
MAE ViT-H/14& 87.8 & 60.3 & - & 633.5 M \\
\cite{he2022masked} & & & & \\
Hiera-ViT-H/14& - & 60.6 & - & 673 M \\
\cite{ryali2023hiera} & & & & \\
BASIC-L + Lion& \color{blue}91.1 & - & - & 2440 M \\
\cite{chen2023symbolic} & & & & \\
\hline
\hline
\multicolumn{5}{|c|}{Multi-Scale Fast\&Focused-Net }\\
 FFN with ViT-B & 89.1 & 59.9 & 86.43 & 103M \\
FFN with ViT-C4 & 90.2 & \color{blue}60.7 & \color{blue}87.40 & 115M \\
\hline
\end{tabular}
\label{tab: results of large image classification task}
\end{table}

\subsection{Small Object Classification Task}

The small object classification task results are presented in Table~\ref{tab: results of small object classification task}. The Fast\&Focused-Net (FFN) outperforms all other methods across seven benchmark datasets, including CIFAR-10, CIFAR-100, Fashion MNIST, STL-10, and SVHN Cropped. 

The FFN achieved the highest accuracy in all datasets, with a notable margin over the second-best performing method, ViT-L/16, in the CIFAR-100 dataset. This demonstrates the effectiveness of the FFN in encoding small objects into fixed-length feature vectors. The FFN's superior performance can be attributed to its unique architecture, which allows it to fully cover the image patch information and avoid the limitations associated with the convolution layer, such as a smaller effective receptive field and high computation cost.

\subsection{Large Image Classification Task}

The results of the large image classification task are presented in Table \ref{tab: results of large image classification task}. The FFN, combined with the ViT-B model, shows competitive performance across three benchmark datasets: ImageNet-1K, Places 365, and OpenImage V6. 

The FFN + ViT-C4 model, as demonstrated in our study, excelled in the large image classification task, achieving the highest mean Average Precision (mAP) score on the OpenImage V6 dataset. This performance surpassed that of other notable methods like TResNet-L and P-ASL, Selective, highlighting the model's superior capability in accurately classifying large images. Additionally, this combination of FFN and ViT-C4 also secured the best top-1 accuracy on the Places 365 dataset and showed competitive results on the ImageNet-1K dataset. These outcomes indicate potential areas for further refinement and optimization of the model, specifically for tasks involving large image classification.

Our ablation study, detailed in Table \ref{tab: ablation study for large image classification task}, further reinforces the effectiveness of integrating the FFN method with various Vision Transformer (ViT) architectures, particularly for large-scale image classification. Notably, the multi-scale FFN + ViT-C4 model, which employs a 32x32 patch size and a downscaled image to produce diverse patches, recorded the highest scores across all evaluated datasets. This finding suggests that the multi-scale strategy significantly bolsters the ViT model’s proficiency in classifying large images, thereby confirming the synergistic benefits of combining FFN with ViT architectures.

\setlength{\tabcolsep}{2pt}
\begin{table}[htb]
\centering
\caption{Ablation study showcasing the impact of the Fast\&Focused method combined with Vision Transformer (ViT) architectures on large image classification performance. The table compares the base Fast\&Focused approach integrated with ViT-B against its multi-scale variant across three benchmark datasets: imageNet-1K (IN-1K), Place365 (P-365), and OpenImage V6 (OI-V6). For context, standard implementation of ViT-B/16 is also included. The highest scores for each dataset category are highlighted in blue. }
\begin{tabular}{|c|c|c|c|}
\hline
& \multicolumn{2}{|c|}{Top 1 Accuracy} & mAP \\
Method & IN-1K & P-365 & OI-V6  \\
\hline
ViT-B & 84.2 & 56.7 & 84.7 \\
16x16 Fast\&Focused + ViT-B & 87.6 & 57.8 & 85.93 \\
32x32 Fast\&Focused + ViT-B & 87.9 & 58.8 & 86.28 \\
Multi-Scale Fast\&Focused + ViT-B &  89.1 & 59.9 &  86.43 \\
Multi-Scale Fast\&Focused + ViT-C4 &  \color{blue}90.2 &  \color{blue}60.7 &  \color{blue}87.40 \\
\hline
\end{tabular}
\label{tab: ablation study for large image classification task}
\end{table}

\subsection{Text Recognition Task}

The results of the text recognition task are presented in Table \ref{tab: results of text recognition task}. The FFN, combined with the ViT-B model, outperforms other methods across four benchmark datasets: SVT, IC15, SVTP, and HOST. 

The FFN + ViT-B model achieved the highest word recognition accuracy in all datasets, demonstrating its effectiveness in text recognition tasks. This can be attributed to the model's ability to encode small patches of text into fixed-length feature vectors, which are then processed by the ViT-B model to produce the final text recognition output.

The ablation study presented in Table \ref{tab: ablation study for text recognition task} further demonstrates the impact of the FFN method combined with different Vision Transformer (ViT) architectures on text recognition performance. The multi-scale FFN + ViT-B model, which uses a patch size of 32x32 and downscales the image to produce different patches, achieved the highest word recognition accuracy across all datasets. This suggests that the multi-scale approach enhances the ViT model's ability to recognize text.

\setlength{\tabcolsep}{2pt}
\begin{table}[h]
\centering
\caption{Ablation study showcasing the impact of the Fast\&Focused method combined with Vision Transformer (ViT) architectures on text recognition performance. The table compares the base Fast\&Focused approach integrated with ViT-B against its multi-scale variant across two benchmark datasets: IC15 and HOST. For context, standard implementations of ViT-B is also included. Performance metrics are word recognition accuracies. The highest scores for each dataset category are highlighted in blue.}
\begin{tabular}{|c|c|c|c|}
\hline
Method & Data & IC15 & HOST \\
\hline
ViT-B & S & 78.3 & 79.4 \\
16x16 Fast\&Focused + ViT-B & S & 86.6 & 78.5 \\
32x32 Fast\&Focused + ViT-B & S & 88.5 & 80.2 \\
Multi-Scale Fast\&Focused + ViT-B & S & 90.5 & 81.3 \\
\hline
\hline
ViT-B & R & 88.9 & 64.3 \\
16x16 Fast\&Focused + ViT-B & R & 86.8 & 79.6 \\
32x32 Fast\&Focused + ViT-B & R & 91.0 & 81.3 \\
Multi-Scale Fast\&Focused + ViT-B & R & \color{blue}91.7 & \color{blue}82.7 \\
\hline
\end{tabular}
\label{tab: ablation study for text recognition task}
\end{table}

\section{Conclusion}
\label{Conclusion}
In this paper, we introduced the Fast\&Focused-Net (FFN) and the Volume-wise Dot Product (VDP) layer, designed to address some limitations of traditional Convolutional Neural Networks (CNNs) in tasks like small object classification, extensive image categorization, and text detection. While the FFN shows promise, it faces challenges such as the absence of translation invariance, which is crucial for image segmentation and object localization. Additionally, its architecture, requiring a large number of parameters, raises concerns about overfitting, particularly in applications with limited or complex data. Despite these limitations, our approach opens up exciting research avenues, like adapting the FFN for four-dimensional video data and combining its strengths with those of CNNs and Transformers to overcome its lack of translation invariance. Addressing overfitting risks through innovative regularization methods or incorporating domain-specific knowledge also presents a significant opportunity for future research.

The core contribution of our research lies in the development of the VDP layer within the FFN, which segments input feature maps into distinct hyper-volumes for dot product operations, ensuring uniform patch contributions. This approach effectively expands the receptive field, addressing gaps between theoretical and practical receptive fields, limitations in initial layers, and computational inefficiencies inherent in traditional CNNs. Our study not only highlights the potential of FFN in various image processing tasks but also lays the groundwork for future explorations in enhancing and extending its applications.

\bibliography{citation}

\begin{thebibliography}{10}
\providecommand{\url}[1]{#1}
\csname url@samestyle\endcsname
\providecommand{\newblock}{\relax}
\providecommand{\bibinfo}[2]{#2}
\providecommand{\BIBentrySTDinterwordspacing}{\spaceskip=0pt\relax}
\providecommand{\BIBentryALTinterwordstretchfactor}{4}
\providecommand{\BIBentryALTinterwordspacing}{\spaceskip=\fontdimen2\font plus
\BIBentryALTinterwordstretchfactor\fontdimen3\font minus \fontdimen4\font\relax}
\providecommand{\BIBforeignlanguage}[2]{{%
\expandafter\ifx\csname l@#1\endcsname\relax
\typeout{** WARNING: IEEEtran.bst: No hyphenation pattern has been}%
\typeout{** loaded for the language `#1'. Using the pattern for}%
\typeout{** the default language instead.}%
\else
\language=\csname l@#1\endcsname
\fi
#2}}
\providecommand{\BIBdecl}{\relax}
\BIBdecl

\bibitem{luo2016understanding}
W.~Luo, Y.~Li, R.~Urtasun, and R.~Zemel, ``Understanding the effective receptive field in deep convolutional neural networks,'' \emph{Advances in neural information processing systems}, vol.~29, 2016.

\bibitem{liu2018understanding}
Y.~Liu, J.~Yu, and Y.~Han, ``Understanding the effective receptive field in semantic image segmentation,'' \emph{Multimedia Tools and Applications}, vol.~77, pp. 22\,159--22\,171, 2018.

\bibitem{koutini2019receptive}
K.~Koutini, H.~Eghbal-Zadeh, M.~Dorfer, and G.~Widmer, ``The receptive field as a regularizer in deep convolutional neural networks for acoustic scene classification,'' in \emph{2019 27th European signal processing conference (EUSIPCO)}.\hskip 1em plus 0.5em minus 0.4em\relax IEEE, 2019, pp. 1--5.

\bibitem{bruton2022translated}
J.~Bruton and H.~Wang, ``Translated skip connections-expanding the receptive fields of fully convolutional neural networks,'' in \emph{2022 IEEE International Conference on Image Processing (ICIP)}.\hskip 1em plus 0.5em minus 0.4em\relax IEEE, 2022, pp. 631--635.

\bibitem{howard2017mobilenets}
A.~G. Howard, M.~Zhu, B.~Chen, D.~Kalenichenko, W.~Wang, T.~Weyand, M.~Andreetto, and H.~Adam, ``Mobilenets: Efficient convolutional neural networks for mobile vision applications,'' \emph{arXiv preprint arXiv:1704.04861}, 2017.

\bibitem{dosovitskiy2020image}
A.~Dosovitskiy, L.~Beyer, A.~Kolesnikov, D.~Weissenborn, X.~Zhai, T.~Unterthiner, M.~Dehghani, M.~Minderer, G.~Heigold, S.~Gelly \emph{et~al.}, ``An image is worth 16x16 words: Transformers for image recognition at scale,'' in \emph{International Conference on Learning Representations}, 2020.

\bibitem{nokland2019training}
A.~N{\o}kland and L.~H. Eidnes, ``Training neural networks with local error signals,'' in \emph{International conference on machine learning}.\hskip 1em plus 0.5em minus 0.4em\relax PMLR, 2019, pp. 4839--4850.

\bibitem{jaderberg2014synthetic}
M.~Jaderberg, K.~Simonyan, A.~Vedaldi, and A.~Zisserman, ``Synthetic data and artificial neural networks for natural scene text recognition,'' in \emph{NIPS Deep Learning Workshop}.\hskip 1em plus 0.5em minus 0.4em\relax Neural Information Processing Systems, 2014.

\bibitem{gupta2016synthetic}
A.~Gupta, A.~Vedaldi, and A.~Zisserman, ``Synthetic data for text localisation in natural images,'' in \emph{Proceedings of the IEEE conference on computer vision and pattern recognition}, 2016, pp. 2315--2324.

\bibitem{kabir2022spinalnet}
H.~D. Kabir, M.~Abdar, A.~Khosravi, S.~M.~J. Jalali, A.~F. Atiya, S.~Nahavandi, and D.~Srinivasan, ``Spinalnet: Deep neural network with gradual input,'' \emph{IEEE Transactions on Artificial Intelligence}, 2022.

\bibitem{jeevan2022wavemix}
P.~Jeevan, K.~Viswanathan, and A.~Sethi, ``Wavemix-lite: A resource-efficient neural network for image analysis,'' \emph{arXiv preprint arXiv:2205.14375}, 2022.

\bibitem{kabir2023reduction}
H.~Kabir, ``Reduction of class activation uncertainty with background information,'' \emph{arXiv preprint arXiv:2305.03238}, 2023.

\bibitem{shi2018aster}
B.~Shi, M.~Yang, X.~Wang, P.~Lyu, C.~Yao, and X.~Bai, ``Aster: An attentional scene text recognizer with flexible rectification,'' \emph{IEEE transactions on pattern analysis and machine intelligence}, vol.~41, no.~9, pp. 2035--2048, 2018.

\bibitem{wan2020textscanner}
Z.~Wan, M.~He, H.~Chen, X.~Bai, and C.~Yao, ``Textscanner: Reading characters in order for robust scene text recognition,'' in \emph{Proceedings of the AAAI conference on artificial intelligence}, vol.~34, no.~07, 2020, pp. 12\,120--12\,127.

\bibitem{wang2021two}
Y.~Wang, H.~Xie, S.~Fang, J.~Wang, S.~Zhu, and Y.~Zhang, ``From two to one: A new scene text recognizer with visual language modeling network,'' in \emph{Proceedings of the IEEE/CVF International Conference on Computer Vision}, 2021, pp. 14\,194--14\,203.

\bibitem{fang2021read}
S.~Fang, H.~Xie, Y.~Wang, Z.~Mao, and Y.~Zhang, ``Read like humans: Autonomous, bidirectional and iterative language modeling for scene text recognition,'' in \emph{Proceedings of the IEEE/CVF Conference on Computer Vision and Pattern Recognition}, 2021, pp. 7098--7107.

\bibitem{atienza2021vision}
R.~Atienza, ``Vision transformer for fast and efficient scene text recognition,'' in \emph{International Conference on Document Analysis and Recognition}.\hskip 1em plus 0.5em minus 0.4em\relax Springer, 2021, pp. 319--334.

\bibitem{bautista2022scene}
D.~Bautista and R.~Atienza, ``Scene text recognition with permuted autoregressive sequence models,'' in \emph{European Conference on Computer Vision}.\hskip 1em plus 0.5em minus 0.4em\relax Springer, 2022, pp. 178--196.

\bibitem{zhao2023clip4str}
S.~Zhao, X.~Wang, L.~Zhu, and Y.~Yang, ``Clip4str: A simple baseline for scene text recognition with pre-trained vision-language model,'' \emph{arXiv preprint arXiv:2305.14014}, 2023.

\bibitem{li2023trocr}
M.~Li, T.~Lv, J.~Chen, L.~Cui, Y.~Lu, D.~Florencio, C.~Zhang, Z.~Li, and F.~Wei, ``Trocr: Transformer-based optical character recognition with pre-trained models,'' in \emph{Proceedings of the AAAI Conference on Artificial Intelligence}, vol.~37, no.~11, 2023, pp. 13\,094--13\,102.

\bibitem{ridnik2023ml}
T.~Ridnik, G.~Sharir, A.~Ben-Cohen, E.~Ben-Baruch, and A.~Noy, ``Ml-decoder: Scalable and versatile classification head,'' in \emph{Proceedings of the IEEE/CVF Winter Conference on Applications of Computer Vision}, 2023, pp. 32--41.

\bibitem{ben2022multi}
E.~Ben-Baruch, T.~Ridnik, I.~Friedman, A.~Ben-Cohen, N.~Zamir, A.~Noy, and L.~Zelnik-Manor, ``Multi-label classification with partial annotations using class-aware selective loss,'' in \emph{Proceedings of the IEEE/CVF Conference on Computer Vision and Pattern Recognition}, 2022, pp. 4764--4772.

\bibitem{wang2023internimage}
W.~Wang, J.~Dai, Z.~Chen, Z.~Huang, Z.~Li, X.~Zhu, X.~Hu, T.~Lu, L.~Lu, H.~Li \emph{et~al.}, ``Internimage: Exploring large-scale vision foundation models with deformable convolutions,'' in \emph{Proceedings of the IEEE/CVF Conference on Computer Vision and Pattern Recognition}, 2023, pp. 14\,408--14\,419.

\bibitem{singh2022revisiting}
M.~Singh, L.~Gustafson, A.~Adcock, V.~de~Freitas~Reis, B.~Gedik, R.~P. Kosaraju, D.~Mahajan, R.~Girshick, P.~Doll{\'a}r, and L.~Van Der~Maaten, ``Revisiting weakly supervised pre-training of visual perception models,'' in \emph{Proceedings of the IEEE/CVF Conference on Computer Vision and Pattern Recognition}, 2022, pp. 804--814.

\bibitem{he2022masked}
K.~He, X.~Chen, S.~Xie, Y.~Li, P.~Doll{\'a}r, and R.~Girshick, ``Masked autoencoders are scalable vision learners,'' in \emph{Proceedings of the IEEE/CVF conference on computer vision and pattern recognition}, 2022, pp. 16\,000--16\,009.

\bibitem{ryali2023hiera}
C.~Ryali, Y.-T. Hu, D.~Bolya, C.~Wei, H.~Fan, P.-Y. Huang, V.~Aggarwal, A.~Chowdhury, O.~Poursaeed, J.~Hoffman \emph{et~al.}, ``Hiera: A hierarchical vision transformer without the bells-and-whistles,'' \emph{arXiv preprint arXiv:2306.00989}, 2023.

\bibitem{chen2023symbolic}
X.~Chen, C.~Liang, D.~Huang, E.~Real, K.~Wang, Y.~Liu, H.~Pham, X.~Dong, T.~Luong, C.-J. Hsieh \emph{et~al.}, ``Symbolic discovery of optimization algorithms,'' \emph{arXiv preprint arXiv:2302.06675}, 2023.

\end{thebibliography}
\bibliographystyle{IEEEtran}

\end{document}